\documentclass[10pt,twocolumn,letterpaper]{article} 

\usepackage{avss}
\usepackage{times}
\usepackage{epsfig}
\usepackage{graphicx}
\usepackage{amsmath}
\usepackage{amssymb}
\usepackage{tabularx}
\usepackage{cite}
\usepackage{array}
\usepackage{caption}
\usepackage{amsmath,amssymb,amsfonts}
\usepackage{algorithmic}
\usepackage{textcomp}
\usepackage{xcolor}

\avssfinalcopy 

\def\avssPaperID{14} 

\ifavssfinal\fi
\begin{document}

\title{Feature Fusion for Human Activity Recognition using Parameter-Optimized Multi-Stage Graph Convolutional Network and Transformer Models}

\newif\ifavssfinal
\avssfinaltrue 

\author{Mohammad Belal\\
Khalifa University of Science and Technology\\
Abu Dhabi, United Arab Emirates\\
{\tt\small 100062548@ku.ac.ae}
\and
Taimur Hassan\\
Abu Dhabi University\\
Abu Dhabi, United Arab Emirates\\
{\tt\small taimur.hassan@adu.ac.ae}
\and
Abdelfatah Ahmed\\
Khalifa University of Science and Technology\\
Abu Dhabi, United Arab Emirates\\
{\tt\small 100059689@ku.ac.ae}
\and
Ahmad Aljarah\\
Khalifa University of Science and Technology\\
Abu Dhabi, United Arab Emirates\\
{\tt\small ahmad.aljarah@ku.ac.ae}
\and
Nael Alsheikh\\
Khalifa University of Science and Technology\\
Abu Dhabi, United Arab Emirates\\
{\tt\small 100062606@ku.ac.ae}
\and
Irfan Hussain\\
Khalifa University of Science and Technology\\
Abu Dhabi, United Arab Emirates\\
{\tt\small irfan.hussain@ku.ac.ae}
}
\maketitle
\thispagestyle{empty}
\let\thefootnote\relax\footnote{979-8-3503-7428-5/24/\$31.00 ©2024 IEEE}
\begin{abstract}
   Human activity recognition is a crucial area of research that involves understanding human movements using computer and machine vision technology. Deep learning has emerged as a powerful tool for this task, with models such as Convolutional Neural Networks (CNNs) and Transformers being employed to capture various aspects of human motion. One of the key contributions of this work is the demonstration of the effectiveness of feature fusion in improving human activity recognition accuracy, which has important implications for the development of more accurate and robust activity recognition systems. This approach addresses a limitation in the field, where the performance of existing models is often limited by their inability to capture both spatial and temporal features effectively. This work presents an approach for human activity recognition using sensory data extracted from four distinct datasets: HuGaDB, PKU-MMD, LARa, and TUG. Two models, the Parameter-Optimized Multi-Stage Graph Convolutional Network (PO-MS-GCN) and a Transformer, were trained and evaluated on each dataset to calculate accuracy and F1-score. Subsequently, the features from the last layer of each model were combined and fed into a classifier. The findings prove that PO-MS-GCN outperforms state-of-the-art models in human activity recognition. Specifically, HuGaDB achieved an accuracy of 92.7\% and f1-score of 95.2\%, TUG achieved an accuracy of 93.2\% and f1-score of 98.3\%, while LARa and PKU-MMD achieved lower accuracies of 64.31\% and 69\%, respectively, with corresponding f1-scores of 40.63\% and 48.16\%. Moreover, feature fusion exceeded the PO-MS-GCN’s results in PKU-MMD, LARa, and TUG datasets.
\end{abstract}
\section{Introduction}

Human activity recognition is a critical task in computer vision, which involves identifying and categorizing human actions within video data. The objective of human activity recognition is to understand and analyze human activities in complex, long videos, with applications ranging from video surveillance to skill assessment \cite{Memon2021-wa}. It plays a vital role in various applications, including automatically classifying videos, and enabling precise and timely adjustments in response to human actions, such as in the case of exoskeletons \cite{Kumar2021-hm}\cite{Fang2020-et}. Recent studies have explored various techniques for human activity recognition, including deep learning-based methods, which have shown significant improvements in the accuracy and efficiency of human activity recognition \cite{De_Miguel-Fernandez2023-rh} \cite{Rahayu2023-ya}.

The integration of deep learning with sensory data has facilitated the development of sophisticated models capable of understanding and analyzing human activities video data. Deep learning models have shown promise in improving the synchronization between human movement and mechanical devices, such as exoskeletons, by enabling precise and timely adjustments in response to human actions. The reduction of the time gap between human action and mechanical device adjustment, without sacrificing precision and quality, has been a key focus area, and deep learning techniques have played a crucial role in addressing this challenge \cite{Singhania2022-vn}\cite{Vrigkas2015-zx}.

\section{Related Work}
Recent studies have explored various techniques for human activity recognition, including those based on advanced network modules, semi-supervised learning, and skeleton-based methods. For instance, Mohsen proposed a gated recurrent unit (GRU) algorithm to classify human activities, achieving significant accuracy \cite{Mohsen2023-dx}, where other researchers have utilized pre-trained models like ResNet50 and ViT to recognizing human activities in daily living, with accuracies reaching 96\% \cite{Surek2023-cw}. Moreover, Huang et al. proposed a network module called Graph-based Temporal Reasoning Module (GTRM) for human activity recognition, demonstrating the application of advanced network modules in this domain \cite{9500219}. Furthermore, recent studies have explored deep learning-based semi-supervised learning for action recognition, emphasizing the potential of deep learning in addressing the complexities of human activity recognition. Another recent study by Filtjens et al. recently proposed a method for skeleton-based action segmentation using Multi-Stage Spatial-Temporal Graph Convolutional Neural Networks (MS-GCN) \cite{9998567}. The proposed method improves the accuracy of action segmentation by leveraging the spatial and temporal dependencies of human actions. The method uses a MS-GCN to extract features from the skeleton data and then applies a temporal convolutional network to recognize the actions..

Moreover, Liu et al. proposed a novel framework for action segmentation, "Diffusion Action Segmentation" leveraging denoising diffusion models to iteratively generate action predictions from random noise with input video features as conditions \cite{Liu2023-mk}. The framework addresses the challenges of temporal activity recognition by enhancing the modeling of three striking characteristics of human actions, including the position prior, the boundary ambiguity, and the relational dependency.


Despite the advancements in human activity recognition using deep learning models, there are still limitations in the field, such as the struggle to capture both spatial and temporal features effectively, which can limit their performance. To address the limitations, the key contributions of the paper are:
\begin{itemize}
\item \textbf{Multi-modal data:} The study leverages sensory data extracted from four distinct datasets, including HuGaDB, PKU-MMD, LARa, and TUG, to train and evaluate two models, the PO-MS-GCN and a Transformer.
\item \textbf{Model comparison:} The paper compares the performance of the PO-MS-GCN and the Transformer on each dataset, providing insights into the strengths and limitations of each model.
\item \textbf{Feature fusion:} The paper demonstrates the effectiveness of feature fusion in improving human activity recognition accuracy, which has important implications for the development of more accurate and robust activity recognition systems.
\item \textbf{Combining strengths of models:} The paper leverages the strengths of the Transformer in capturing long-range dependencies and temporal patterns, and the PO-MS-GCN in capturing fine-grained spatial and temporal features, demonstrating the potential of combining different models for improved recognition accuracy.
\end{itemize}

\section{Proposed Method}
In this section, we provide an overview of the proposed methodology. Four distinct datasets, namely HuGaDB, PKU-MMD, LARa, and TUG, were leveraged to execute two distinct models: Parameter-Optimized Multi-Stage Graph Convolutional Network (PO-MS-GCN) and a Transformer for the purpose of recognizing human actions. Subsequently, the features from the last layer of each model were extracted and combined through concatenation, followed by the transmission of the combined features to a Fully Connected Network classifier.

\begin{figure*}[ht]
    \centering
    \includegraphics[width=1\textwidth]{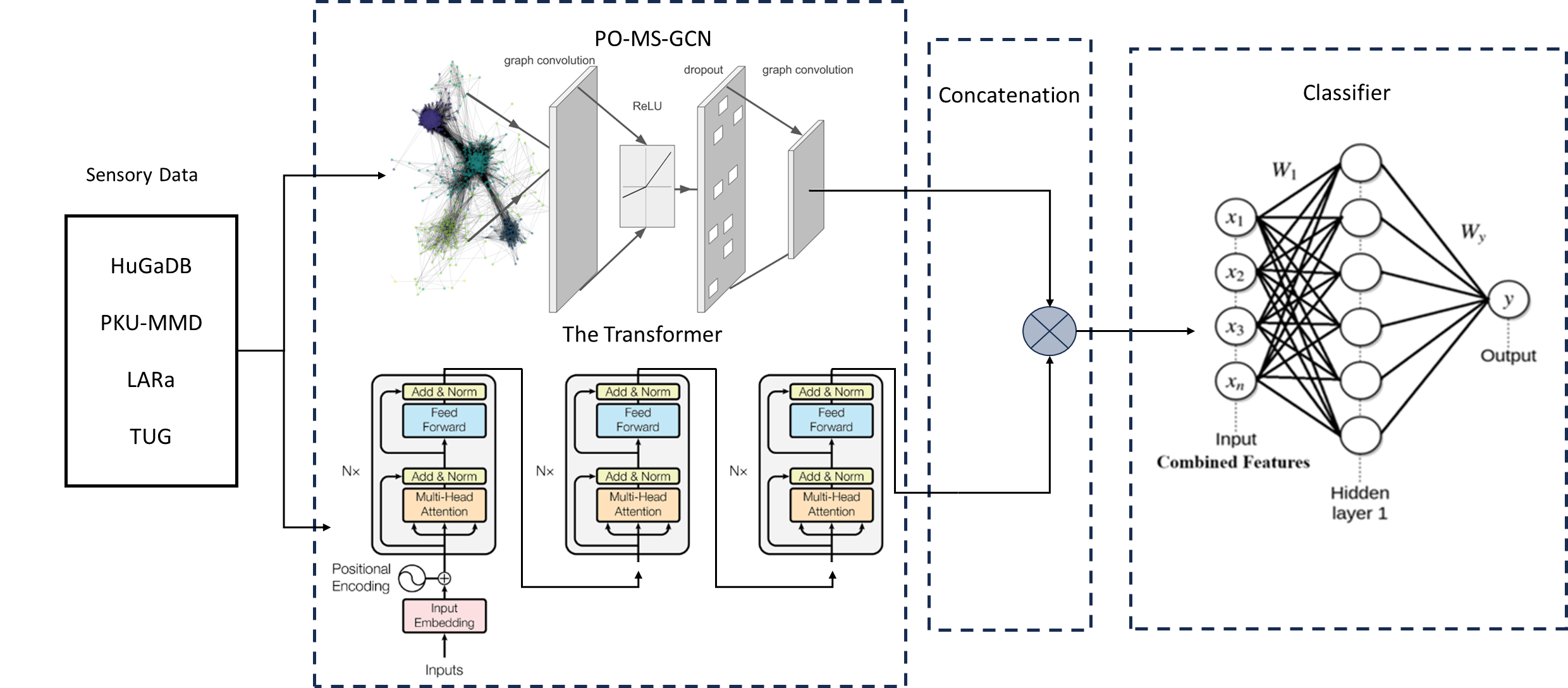}
    \caption{The features were gathered from the (PO-MS-GCN) and the Transformer, then both features were combined through concatenation. The combined features were then passed to the classifier to be used as an input.}
    \label{figure_1}
\end{figure*}

\subsection{Parameter-Optimized Multi-Stage Graph Convolutional Network (PO-MS-GCN)}
The Parameter-Optimized MS-GCN model, tailored for skeleton-based activity recognition tasks, has demonstrated promising outcomes in various experiments. The model's architecture was refined through better parameter tuning compared to the MS-GCN \cite{9998567}, resulting in improved performance. Rooted in a spatiotemporal graph representation of human skeletal movements, PO-MS-GCN effectively captures spatial and temporal dependencies within the input sequence. Comprising multiple stages, each housing its convolutional graph network (GCN) layers, PO-MS-GCN model effectively captures and propagates information across the spatiotemporal graph representation of human skeletal movements. By processing a sequence of human skeletal movements represented as a spatiotemporal graph, the model conducts spatial and temporal graph convolutions in each stage, followed by a graph pooling operation to reduce the graph size. The resulting feature maps are subsequently forwarded to the next stage for further processing. In the training of the MS-GCN model for activity recognition, two loss functions were employed: the Cross-Entropy loss (CE) and the Mean Squared Error (MSE) loss. The CE loss is defined as:
\begin{equation}
\begin{aligned}
& C E=\sum_{s t=1}^{s t} C E_{s t, c l s}
\end{aligned}
\end{equation}
\\ \begin{equation}
\begin{aligned}
& C E_{c l s}=\frac{-1}{N} \sum_n y_{n, l} \log \left(\hat{y}_{n, l}\right),
\end{aligned}
\end{equation}
Where the total loss over all stages (st) is represented by CE, while $CE_{\text{cls}}$
denotes the loss between the ground truth label $y_{\text{n,l}}$ and the predicted probability label $\hat{y}_{\text{n,l}}$ for class $l$ at sample $n$, respectively. At the same time, the MSE loss is commonly used for regression tasks. The use of these two loss functions is aimed at improving its performance in activity recognition by considering both classification and regression objectives was proposed by \cite{Farha2019-ke}. The MSE loss is defined as:
\begin{equation}
M S E=\frac{1}{N} \sum_{i=1}^N\left(y_i-\hat{y}_i\right)^2
\end{equation}
Where N denotes the number of samples and $y_{\text{i}}$ is the true probability, while $\hat{y}_{\text{i}}$ is the predicted probability for a sample $i$ in $N$ samples. The combination of the MSE loss and CE loss led to the following equation:
\begin{equation}
L_{\text {combined }}=\sum_{s t=1}^{S t} C E+\lambda M S E \text {, }
\end{equation}
Where $L_{\text {combined }}$ denotes the combined loss for $\mathrm{CE}$ and $\mathrm{MAE}$, while $\lambda$ determines the weight of the MSE loss in the combined loss, the combined loss function was proposed to mitigate over-recognition errors arising from excessively high sample frequency in predictions. The model's final output consists of action labels corresponding to each frame in the input sequence, rendering it suitable for diverse activity recognition tasks.

\subsection{Transformer model}
A Transformer model was proposed for the purpose of human activity recognition. The proposed model was trained using the same hyperparameters previously applied to train PO-MS-GCN. The input to the Transformer model is not explicitly tokenized. Instead, the input data, which represents the skeletal information, is directly fed into the Transformer model. The input data is expected to be a 3D tensor with shape (batch\_size, seq\_len, input\_size), where seq\_len is the sequence length and input\_size is the number of features for each token in the sequence. The Transformer model is working based on the self-attention mechanism, which allows it to weigh the importance of each token in the sequence relative to every other token. This is done without explicit positional embeddings.

In the self-attention mechanism, each token in the input sequence is associated with a weight that determines how much attention should be paid to that token when generating the output for another token. These weights are learned during the training process and depend on the content of the tokens themselves, not their positions in the sequence. The benefits of utilizing the transformer architecture include enhanced feature extraction, better representation of long-range dependencies. This approach could lead to more accurate and efficient activity recognition.

\subsection{Feature Fusion}
Feature fusion is a technique used in machine learning and deep learning to combine features extracted from multiple models to improve the overall performance of a model \cite{6918213}. In this study, two models, Parameter-Optimized MS-GCN and the Transformer, were used to extract features from their respective last layers \cite{Zhang2021-io} \cite{Mungoli2023-hl}. The features extracted from both models were then combined using concatenation, which allows the diverse and complementary information captured by each model to be integrated, potentially enhancing the overall representation and predictive capabilities. Feature fusion is an essential technique in machine learning and deep learning, and it has been widely studied in the literature. Researchers have proposed various feature fusion methods, such as guided training and attentional feature fusion, to improve the performance of classification tasks \cite{Dai_undated-ji}.

\subsection{The Fully Connected Classifier}
The fully connected network serves as a classifier in this study, where the classifier takes the combined features as input \cite{Basha2019-ji} \cite{Kalayci2022-cs}. The network consists of a Batch normalization layer, two dense layers where one of them is an output layer, and a flatten layer. The Adaptive Moment Estimation (Adam), which is a popular algorithm used in deep learning that helps adjust the parameters of a neural network in real-time to improve its accuracy and speed \cite{Kingma2014-at}. The Adam optimizer is known for its adaptive learning rate and momentum-based approach, which can help the neural network learn faster and converge more quickly towards the optimal set of parameters that minimize the cost or loss function. CE loss served as a loss function in the classifier utilized in this study. Figure \ref{figure_1} shows an overview of the system created in this study.

\section{Experimental Setup}
\subsection{Datasets}
\subsubsection{HuGaDB}
A human gait database designed to facilitate research in gait analysis and recognition, encompasses gait data obtained from individuals with diverse walking abilities and under varying conditions \cite{Chereshnev2018-nb}. The database comprises 12 distinct gait actions, each accompanied by labels for the stance, swing, and double support phases, as well as details regarding the walking surface, footwear, and the presence of assistive devices. Data collection involved the use of various sensors, including force plates, inertial measurement units, and electromyography sensors.

\subsubsection{PKU-MMD}
The PKU-MMD motion capture dataset is a comprehensive resource for researchers in 3D human motion analysis \cite{Liu2017-pn}. The dataset contains over 2000 motion sequences performed by 78 subjects engaging in various daily and sports activities, including walking, jogging, cycling, and dancing. The motion capture data was obtained using an optical motion capture system comprising 12 high-speed cameras, which captured the motion of reflective markers placed on the subject’s body.

\subsubsection{LARa}
The Logistic Activity Recognition Challenge (LARa) database, a publicly accessible dataset tailored for the assessment of activity recognition algorithms in logistic tasks, encompasses data captured from four sensors, including a tri-axial accelerometer, gyroscope, magnetometer, and barometer. This dataset encompasses six activities associated with logistic tasks, such as stacking and moving boxes, and involves the participation of 20 individuals \cite{Niemann2020-rf}.

\subsubsection{TUG}
The Timed Up and Go (TUG) dataset, as described in "The timed "Up \& Go": a test of basic functional mobility for frail elderly persons," serves as a valuable resource for researchers focusing on mobility, balance, and fall risk assessment, particularly in elderly individuals and those with neurological conditions such as stroke and Parkinson's \cite{Podsiadlo1991-tc}. This dataset encompasses sensor-based data obtained from patients performing the TUG test, a widely used clinical assessment for evaluating mobility and fall risk. The dataset includes data from healthy individuals and those with neurological conditions, offering a comprehensive view of mobility and balance in diverse populations. With various sensor modalities, including accelerometers, gyroscopes, and magnetometers.
\subsection{Evaluation Metrics}
The performance of the proposed model was evaluated using several standard evaluation metrics, including overall accuracy and F1-score. The F1-score is used as a segment-wise evaluation metric \cite{Farha2019-ke}, where the predicted action segment is classified as either true positive (TP) or false positive (FP), as reported by \cite{Lea2017-re}. The segment's intersection over union (IoU) was compared to that of the corresponding expert annotation. If the IoU crosses a predetermined overlap threshold, it is classified as a true positive segment (TP); otherwise, it is classified as a false positive segment (FP). The F1 score is defined as the following:
\begin{equation}
F1-\text {score}=\frac{T P}{0.5(F N+F P)+T P}
\end{equation}

Accuracy is used as a sample-wise evaluation metric \cite{Farha2019-ke}, which measures the proportion of correctly classified instances. It is a useful metric for evaluating classification tasks when the classes are balanced, meaning they have roughly equal samples. Accuracy is easy to interpret and provides a straightforward measure of a model's performance. The accuracy equation is defined as:
\begin{equation}
Acc =\frac{T P+T N}{F N+F P+T N+T P}
\end{equation}

These metrics were calculated at the segment level for each action class. The reported evaluation metrics were obtained by averaging the performance overall cross-validation folds. The results were reported using both per-frame and per-action evaluation protocols. The per-action evaluation protocol evaluated the model's performance on different action classes.

\begin{table}[b]
\raggedleft
\caption{The F1-score at 50 Hz and Accuracy results for PO-MS-GCN and the Transformer activity recognition models on PKU-MMD, HuGaDB, LARa, and TUG datasets.}
\captionsetup{skip=1ex}
\begin{tabularx}{1\linewidth}{|>{\centering\arraybackslash}p{1.15cm}|>{\centering\arraybackslash}p{1.55cm}|>{\centering\arraybackslash}p{0.95cm}|>{\centering\arraybackslash}p{1.55cm}|>{\centering\arraybackslash}p{0.95cm}|}
\hline \textbf{Model} & \multicolumn{2}{|c|}{ \textbf{PO-MS-GCN} } & \multicolumn{2}{c|}{ \textbf{Transformer} } \\
\hline \textbf{Dataset} & \textbf{Accuracy\%} & \textbf{F1-score\%} & \textbf{Accuracy\%} & \textbf{F1-score\%}\\
\hline HuGaDB & 92.70 & 95.2 & 90.3 & 94 \\
\hline LARa & 64.31 & 40.63 & 59.30 & 30.5 \\
\hline PKU-MMD & 69.0 & 48.16 & 68.3 & 52.9 \\
\hline TUG & 93.20 & 98.3 & 90.9 & 98.1 \\
\hline
\end{tabularx}
\label{table1}
\end{table}

\section{Experimental Analysis and Results}
The aim of the study was to compare the results of the Parameter-Optimized MS-GCN with the results reported by \cite{9998567} and the Transformer model. The goal was to evaluate the reproducibility of the approach and contribute additional insights into the performance of the proposed method to the research community. By comparing the results of different models, the study aimed to provide a better understanding of the strengths and weaknesses of each approach. As shown in Table \ref{table1}, the results obtained for the accuracy and F1 score for the four datasets were obtained by running the code for 100 epochs with a batch size of 4 and a learning rate of 0.0005 while accounting for the sampling factor difference based on the sampling rate in Hertz (Hz) for each dataset.


\begin{table}[h]
\centering
\caption{Comparison between results reported by \cite{9998567} for the MS-GCN, PO-MS-GCN, and other activity recognition models.}
\captionsetup{skip=1ex}
\begin{tabularx}{0.996\linewidth}
{|>{\centering\arraybackslash}p{2.325cm}|>{\centering\arraybackslash}p{2.325cm}|>{\centering\arraybackslash}p{2.325cm}|}
\hline 
\textbf{HuGaDB} & \textbf{Accuracy\%} & \textbf{F1-score\%}\\ 
\hline
TCN & 88.3 & 56.8\\
Bi-LSTM & 86.1 & 81.5\\
MS-GCN & 90.4 & 93.0\\
ST-GCN & 88.7 & 67.7\\
MS-TCN & 86.8 & 89.9\\
Transformer & 90.3 & 94.0\\
PO-MS-GCN & \textbf{92.7} & \textbf{95.2}\\
\hline
\textbf{LARa} & \textbf{Accuracy\%} & \textbf{F1-score\%}\\ 
\hline
TCN & 61.5 & 20.0\\
Bi-LSTM & 63.9 & 32.3\\
MS-GCN & 65.6 & \textbf{43.6}\\
ST-GCN & \textbf{67.9} & 25.8\\
MS-TCN & 65.8 & 39.6\\
Transformer & 59.3 & 30.5\\
PO-MS-GCN & 64.31 & 40.63\\
\hline
\textbf{PKU-MMD} & \textbf{Accuracy\%} & \textbf{F1-score\%}\\
\hline
TCN & 61.9 & 13.8\\
Bi-LSTM & 59.6 & 22.7\\
MS-GCN & 68.5 & 51.6\\
ST-GCN & 64.9 & 15.5\\
MS-TCN & 65.5 & 46.3\\
Transformer & 68.3 & \textbf{52.9}\\
PO-MS-GCN & \textbf{69.0} & 48.16\\
\hline
\textbf{TUG} & \textbf{Accuracy\%} & \textbf{F1-score\%}\\ 
\hline
TCN & 92.7 & 84.4\\
Bi-LSTM & 93.2 & 97.1\\
MS-GCN & \textbf{93.6} & 97.9\\
ST-GCN & 93.2 & 93.8\\
MS-TCN & 92.7 & 96.5\\
Transformer & 90.9 & \textbf{98.1}\\
PO-MS-GCN & 93.2 & \textbf{98.3}\\
\hline
\end{tabularx}
\label{table2}
\end{table}

\begin{table}[h]
\raggedleft
\caption{Results acquired from feature fusion using a combination of features from the PO-MS-GCN and the Transformer.}
\captionsetup{skip=1ex}
\begin{tabularx}{1\linewidth}
{|m{2cm}|m{2.5cm}|m{2.5cm}|m{2.5cm}|m{2.5cm}|}
\hline \textbf{Technique} & \multicolumn{2}{|c|}{ \textbf{Feature Fusion}} \\
\hline \textbf{Dataset} & \textbf{Accuracy\%} & \textbf{F1-score\%}\\
\hline HuGaDB  & 84.70 & 88.20 \\
\hline LARa  & 59.30 & 50.48 \\
\hline PKU-MMD  & 96.61 & 94.95 \\
\hline TUG  & 98.44 & 97.66 \\
\hline
\end{tabularx}
\label{table_3}
\end{table}

Table \ref{table2} demonstrates a discrepancy in performance between the results reported by \cite{9998567}, the results observed from the proposed PO-MS-GCN for the four datasets, the results gathered from the Transformer, and the results from other activity recognition models. The proposed PO-MS-GCN showed an increase of 2.3\% in accuracy and 2.2\% in F1-score compared to the MS-GCN for HuGaDB and a decrease of 3.59\% in accuracy compared to ST-GCN and a decrease of 2.97\% in F1-score compared to MS-GCN for LARa dataset. Conversely, for PKU-MMD, the proposed PO-MS-GCN showed a slight improvement in accuracy of 0.5\% compared to the MS-GCN but had an F1-score value of 3.44\% less, while the Transformer achieved the best F1-score with 52.9\%. Furthermore, the F1-score achieved using the proposed PO-MS-GCN for the TUG dataset surpassed MS-GCN's by 0.4\%. The proposed PO-MS-GCN's accuracy was less than the result reported for the MS-GCN by 0.44\%. For the Transformer model, it can be seen that it has achieved close results to the ones reported by \cite{9998567} for HuGaDB, but not for the PKU-MMD dataset, where it had much lower accuracy compared to PO-MS-GCN and MS-GCN. For the TUG dataset, it was found that it had the same F1-score value but lower accuracy than the other two models.

Table \ref{table_3} highlights the main findings of the feature fusion method for the PO-MS-GCN and the Transformer in the context of human activity recognition. The use of feature fusion demonstrated adaptability in this application, leading to improvements in accuracy and f1-score. For the LARa dataset, feature fusion resulted in approximately 10\% improvement in f1-score. A significant difference was observed between the results obtained using feature fusion and the PO-MS-GCN alone for the PKU-MMD dataset, with 27.6\% and 46.8\% improvements in accuracy and f1-score, respectively. Additionally, for the TUG dataset, feature fusion led to an improvement of 5.24\% and 4.84\% in accuracy compared to the PO-MS-GCN and the MS-GCN, respectively. The diversity in the results can be attributed to the diversity of sensors and the quality of sensors utilized in each dataset, which affects the precision of activity recognition.

\section{Conclusion}
This paper demonstrates an effective approach for human activity recognition using sensory data from four distinct datasets, namely: HuGaDB, PKU-MMD, LARa, and TUG. Two models, the Parameter-Optimized Multi-Stage Convolutional Neural Network (PO-MS-GCN) and a Transformer, were trained and evaluated on each dataset, and their performance was assessed using accuracy and F1-score metrics. Feature fusion method was applied on both models to leverage the advantages of the graph convolutional networks and the transformers, where the features from the last layer in both models were combined and passed into a fully connected classifier. The findings indicate that the PO-MS-GCN outperforms state-of-the-art models in human activity recognition. Additionally, the study shows that feature fusion is beneficial for recognizing human activities, surpassing the performance of the PO-MS-GCN in three of the datasets. This approach leverages the strengths of the Transformer in capturing long-range dependencies and temporal patterns, and the PO-MS-GCN in capturing fine-grained spatial and temporal features. Diversity in the results can be noticed accross the datasets which can be attributed to the diversity of sensors and the quality of sensors utilized in each dataset, which affects the precision of activity recognition.


\end{document}